\documentclass[11pt]{article}
\usepackage{coling2020}
\usepackage{times}
\usepackage{url}
\usepackage{latexsym}

\usepackage{microtype}
\usepackage{multirow}
\usepackage{graphicx}
\usepackage{booktabs}
\usepackage{tikz}
\usepackage{pgfplots}
\usepackage{amsmath}
\usepackage{amssymb}
\usepackage{algorithm}
\usepackage{algpseudocode}
\usepackage{caption}
\usepackage{wrapfig}
\usepackage{subfigure}

\pgfplotsset{compat=1.16}

\usetikzlibrary{plotmarks}
\usetikzlibrary{arrows}
\usetikzlibrary{decorations}
\usetikzlibrary{decorations.pathreplacing}
\usetikzlibrary{decorations.markings}
\usetikzlibrary{backgrounds}
\usetikzlibrary{fit}
\usetikzlibrary{positioning}
\usetikzlibrary{calc}
\usetikzlibrary{patterns}
\usetikzlibrary{shadows}
\usetikzlibrary[shapes.multipart]
\usetikzlibrary{pgfplots.groupplots}

\definecolor{ugreen}{rgb}{0,0.5,0}
\definecolor{lyyblue}{RGB}{31,120,180}
\definecolor{lyygreen}{RGB}{51,160,44}
\definecolor{lyyred}{RGB}{227,26,28}

\newcommand{\vecmap}{VecMap}

\newcommand{\tab}[1]{Table \ref{#1}}%
\newcommand{\fig}[1]{Figure \ref{#1}}%
\newcommand{\alg}[1]{Algorithm \ref{#1}}%

\colingfinalcopy 

\title{A Simple and Effective Approach to Robust Unsupervised Bilingual Dictionary Induction}

\author{Yanyang Li$^1$\thanks{\ \ Authors contributed equally.}\ , Yingfeng Luo$^1$\footnotemark[1]\ , Ye Lin$^1$, Quan Du$^1$,\\
{\bf Huizhen Wang$^1$, Shujian Huang$^{3,4}$, Tong Xiao$^{1,2}$\thanks{\ \ Corresponding author.} \and Jingbo Zhu$^{1,2}$} \\
$^1$Natural Language Processing Lab., Northeastern University, Shenyang, China\\
$^2$NiuTrans Research, Shenyang, China\\
$^3$National Key Laboratory for Novel Software Technology, Nanjing, China\\
$^4$Nanjing University, Nanjing, China\\
{\tt blamedrlee@outlook.com, 1971646@stu.neu.edu.cn},\\
{\tt \{linye2015, duquanneu\}@outlook.com, huangsj@nju.edu.cn},\\
{\tt \{wanghuizhen, xiaotong, zhujingbo\}@mail.neu.edu.cn} \\}

\date{}

\begin{document}
\maketitle
\begin{abstract}
  Unsupervised Bilingual Dictionary Induction methods based on the initialization and the self-learning have achieved great success in similar language pairs, e.g., English-Spanish. But they still fail and have an accuracy of 0\% in many distant language pairs, e.g., English-Japanese. In this work, we show that this failure results from the gap between the actual initialization performance and the minimum initialization performance for the self-learning to succeed. We propose \emph{Iterative Dimension Reduction} to bridge this gap. Our experiments show that this simple method does not hamper the performance of similar language pairs and achieves an accuracy of 13.64$\sim$55.53\% between English and four distant languages, i.e., Chinese, Japanese, Vietnamese and Thai.
\end{abstract}

\section{Introduction}

\blfootnote{
    %
    %
    %
    %
    \hspace{-0.65cm}  
    This work is licensed under a Creative Commons 
    Attribution 4.0 International Licence.
    Licence details:
    \url{http://creativecommons.org/licenses/by/4.0/}.
    
    
}

  Unsupervised Bilingual Dictionary Induction (UBDI) is a task that aims to find the word translations given the monolingual word embeddings of two languages. Recent UBDI methods have shown promising results on similar language pairs such as English-Spanish \cite{DBLP:conf/acl/ArtetxeLA17,DBLP:conf/iclr/LampleCRDJ18,DBLP:conf/naacl/ZhouMWN19,DBLP:conf/nips/HartmannKS19,DBLP:conf/acl/RenLZM20}. These methods are mostly based on the initialization and the self-learning. The initialization first constructs a dictionary from the word embeddings, then the self-learning starts with this dictionary and alternates between refining the source-target word embedding mapping and inducing a new dictionary with this mapping.

  Despite the success of UBDI, recent work has questioned the robustness of UBDI methods on distant language pairs \cite{DBLP:conf/acl/SogaardVR18,vulic-etal-2019-really,DBLP:conf/acl/GlavasLRV19}, e.g., English-Japanese. They show that even for the most robust system \vecmap{} \cite{DBLP:conf/acl/AgirreLA18}, it still fails and has an accuracy of 0\% on 87 out of 210 distant language pairs \cite{vulic-etal-2019-really}. To be consistent with \newcite{DBLP:conf/acl/AgirreLA18}, we define a system `succeeds' when it has an accuracy above 5\% and `fails' otherwise.

  Previous work has investigated how different properties of languages have an impact on UBDI performance \cite{DBLP:conf/acl/SogaardVR18}. In this paper, we take a step further to inspect which part of \vecmap{} breaks down. With a novel similarity metric to evaluate the initialization performance, we observe a gap between the actual initialization performance and the minimum initialization performance for the self-learning to succeed in distant language pairs.

  We find that the dimension reduction approach is very effective in bridging this gap. Therefore, we propose \emph{Iterative Dimension Reduction} (IDR) to improve the robustness of \vecmap{} and avoid performance loss due to dimension reduction. IDR first reduces the dimension of word embeddings and performs unsupervised learning on them. Then it initializes the self-learning on larger dimension embeddings using this learned system. This simple dimension reduction removes unimportant or noisy features, making the algorithm easier to find a proper solution to distant language pairs.

  We evaluate our approach on four similar European language pairs, including English-\{Spanish, French, Italian, German\} (En-\{Es, Fr, It, De\}), and four distant language pairs, including English-\{Chinese, Japanese, Vietnamese, Thai\} (En-\{Zh, Ja, Vi, Th\}). Our method not only has a close performance to the \vecmap{} baseline in similar language pairs but also succeeds in all distant language pairs. In four distant language pairs, our method has an accuracy of 13.64$\sim$55.53\%, whereas the \vecmap{} baseline has an accuracy of 0\% in most cases, as shown in the first row of \tab{tab:semi-supervised}.

\section{The VecMap Method}

  \vecmap{} \cite{DBLP:conf/acl/AgirreLA18} learns weights $W_X$ and $W_Y$ for the source and target word embeddings $X$ and $Y$ and maps them to the same space for inducing the dictionary. It consists of two components:
  \begin{itemize}
    \item \emph{Initialization}. The initialization first computes $M_X=XX^T$ and $M_Y=YY^T$. Then each row of $M_X$ and $M_Y$ is sorted and an initial dictionary $D$ is induced by searching for nearest neighbors between the rows of $\sqrt{M_X}$ and $\sqrt{M_Y}$.
    \item \emph{Self-learning}. With the initial dictionary $D$, the self-learning iterates the following two steps:
    \begin{itemize}
      \item It finds $W_X$ and $W_Y$ that maximize $\sum_i\sum_jD_{ij}((X_{i*}W_X)\cdot(Y_{j*}W_Y))$ for the current dictionary $D$, where $D_{ij}=1$ if the $j$-th target word is the translation of the $i$-th source word and 0 otherwise. An optimal solution is given by $W_X=U$ and $W_Y=V$, where $USV^T=X^TDY$ is the singular value decomposition of $X^TDY$;
      \item A new dictionary $D$ is induced by using the CSLS retrieval \cite{DBLP:conf/iclr/LampleCRDJ18} to extract nearest neighbors in the similarity matrix $P=XW_XW_Y^TY^T$. To avoid being trapped in a poor local optimum and encourage the exploration of possible word translations, similarity scores in $P$ are kept with a probability $p$ and set to 0 otherwise.
    \end{itemize}
  \end{itemize}

  There are some pre-processing and post-processing steps that are crucial to \vecmap{}:
  \begin{itemize}
    \item \emph{Normalization and mean centering} is applied before the initialization and the self-learning. It will normalize all vectors in $X$ and $Y$ to have a unit Euclidean norm. Then these vectors are mean centered dimension-wise and length-normalized again.
    \item \emph{Whitening} \cite{bell1997the} is applied before the last iteration of the self-learning. It transforms $X$ and $Y$ such that each dimension has unit variance and that the dimensions are uncorrelated.
    \item \emph{Symmetric re-weighting} is applied to the mapped embeddings $XW_X$ and $YW_Y$ at the last iteration. It further improves $XW_X$ and $YW_Y$ by weighting dimensions by their cross-correlation $\sqrt{S}$, where $S$ is a diagonal matrix with singular values on its diagonal entries.
    \item \emph{Dewhitening} is the reverse of whitening and applied after symmetric re-weighting if whitening is applied. It restores the variance information.
  \end{itemize}


  \begin{table}[t]
    \centering
    \begin{tabular}{l|cccccccc}
      \toprule
      \multicolumn{1}{c|}{Entry} & En-Zh & Zh-En & En-Ja & Ja-En & En-Vi & Vi-En & En-Th & Th-En \\
      \midrule
      \vecmap{} Accuracy [\%] & 0.07 & 0 & 1.03 & 32.67 & 0.73 & 0.73 & 0 & 0.07 \\
      \midrule
      Maximum Accuracy [\%] & 37.07 & 35.20 & 49.01 & 33.36 & 47.13 & 57.80 & 24.20 & 17.76 \\
      Minimum Dictionary Size & 150 & 74 & 74 & 97 & 137 & 110 & 147 & 154 \\
      \bottomrule
    \end{tabular}
    \caption{\vecmap{} accuracy, the maximum accuracy of using a seed dictionary to initialize the self-learning and the minimum seed dictionary size to obtain that accuracy (we start with 10 pairs to estimate this size and add 10 pairs if the maximum accuracy is not achieved; Results are averaged over 3 runs).}
    \label{tab:semi-supervised}
  \end{table}

\section{When does Unsupervised Learning Fail?}
\label{sec:analysis}

\begin{figure*}[t]
  \begin{minipage}[t]{.49\columnwidth}
    \hspace*{\fill}
    \begin{tikzpicture}
      \begin{axis}[
        width=0.5\linewidth,height=5cm,
        scaled ticks=false,
        xticklabel style={
          /pgf/number format/fixed,
          /pgf/number format/precision=2
        },
        enlargelimits=0.05,
        ymin=0,ymax=100,
        xmin=-0.04,xmax=0.02,
        ylabel={Accuracy [\%]},
        ylabel near ticks,
        xlabel={Dictionary Similarity},
        xlabel near ticks,
        xmajorgrids=true,
        ymajorgrids=true,
        grid style=dashed,
        every tick label/.append style={font=\small},
        label style={font=\small},
        legend style={font=\small,at={(0.5,1)},anchor=south,inner sep=3pt,draw=none},
        legend image post style={scale=0.8},
        legend columns=2,
        legend cell align={left},
      ]
      \addplot [lyyblue,mark=*] coordinates {
        (-0.0024,0.67) (0.0017,0.4) (0.0036,82.2) (0.0099,82.33) (0.0141,82.13)
      };\addlegendentry{En-Es}
      \addplot [lyyred,mark=square*] coordinates {
        (-0.0389,0.60) (-0.0277,0) (-0.0186,82.4) (-0.0134,82.4) (-0.0098,82.2)
      };\addlegendentry{En-Fr}
      \addplot [lyygreen,mark=triangle*] coordinates {
        (-0.0388,0.4) (-0.0317,0.33) (-0.0251,79.13) (-0.0178,79) (-0.0125,79)
      };\addlegendentry{En-It}
      \addplot [orange,mark=diamond*] coordinates {
        (-0.0093,0.33) (-0.0067,1.47) (0.0018,75.2) (0.0063,75.07) (0.0116,75.07)
      };\addlegendentry{En-De}
      \end{axis}
    \end{tikzpicture}
    \hfill
    \begin{tikzpicture}
      \begin{axis}[
        width=0.5\linewidth,height=5cm,
        enlargelimits=0.05,
        ymin=0,ymax=60,
        xmin=0.1,xmax=0.45,
        xlabel={Dictionary Similarity},
        xlabel near ticks,
        xmajorgrids=true,
        ymajorgrids=true,
        grid style=dashed,
        every tick label/.append style={font=\small},
        label style={font=\small},
        legend style={font=\small,at={(0.5,1)},anchor=south,inner sep=3pt,draw=none},
        legend image post style={scale=0.8},
        legend columns=2,
        legend cell align={left},
      ]
      \addplot [lyyblue,mark=*] coordinates {
        (0.3888,0) (0.4186,0) (0.4298,39.67) (0.4393,37.4) (0.449,37.27)
      };\addlegendentry{En-Zh}
      \addplot [lyyred,mark=square*] coordinates {
        (0.378,3.84) (0.3922,0.34) (0.4012,47.43) (0.4125,48.39) (0.4237,48.53)
      };\addlegendentry{En-Ja}
      \addplot [lyygreen,mark=triangle*] coordinates {
        (0.1208,0.13) (0.134,0.53) (0.1478,48) (0.1627,48) (0.1748,47.2)
      };\addlegendentry{En-Vi}
      \addplot [orange,mark=diamond*] coordinates {
        (0.1625,0) (0.1745,0) (0.1865,20.33) (0.1968,21.33) (0.2087,21.67)
      };\addlegendentry{En-Th}
      \end{axis}
    \end{tikzpicture}
    \hspace*{\fill}
    \caption{The accuracy obtained by the self-learning starting from a given dictionary (Accuracy) vs. the dictionary similarity of that starting dictionary (Dictionary Similarity).}\label{fig:acc-vs-sim}
  \end{minipage}
  \hfill
  \begin{minipage}[t]{.49\columnwidth}
    \centering
    \begin{tikzpicture}
      \begin{axis}[
        width=\linewidth,height=5cm,
        symbolic x coords={En-Es,En-Fr,En-It,En-De,En-Zh,En-Ja,En-Vi,En-Th},
        enlarge x limits=0.1,
        enlarge y limits={upper,value=0.35},
        ylabel={Dictionary Similarity $+ 1$},
        ylabel near ticks,
        ybar=0pt,
        xtick=data,
        ytick=\empty,
        nodes near coords,
        every node near coord/.append style={rotate=90,anchor=west,font=\scriptsize},
        every tick label/.append style={font=\footnotesize},
        xticklabel style={rotate=45,anchor=north east,font=\footnotesize,inner sep=0pt,outer sep=2pt},
        bar width=5.5pt,
        xmajorgrids=true,
        ymajorgrids=true,
        grid style=dashed,
        label style={font=\small},
        legend image code/.code={
            \draw [#1] (0cm,-0.1cm) rectangle (0.3cm,0.1cm);
        },
        legend style={font=\small,at={(0.02,0.98)},anchor=north west,inner sep=3pt,column sep=3pt},
        legend cell align={left},
        legend columns=1,
      ]
        \addplot [draw=lyyblue!80,fill=lyyblue!60,pattern=north west lines,pattern color=lyyblue] coordinates {(En-Es,1.0036) (En-Fr,0.98659) (En-It,0.9749) (En-De,1.0018) (En-Zh,1.4298) (En-Ja,1.4125) (En-Vi,1.1478) (En-Th,1.1865)};\addlegendentry{Threshold}
        \addplot [draw=lyygreen!80,fill=lyygreen!60,pattern=north east lines,pattern color=lyygreen] coordinates {(En-Es,1.0955) (En-Fr,1.106) (En-It,1.0935) (En-De,1.09) (En-Zh,1.3889) (En-Ja,1.3907) (En-Vi,1.107) (En-Th,1.1401)};\addlegendentry{Actual}
      \end{axis}
    \end{tikzpicture}
    \captionof{figure}{The minimum initial dictionary similarity to succeed (Threshold) vs. the actual initial dictionary similarity (Actual).}\label{fig:required-vs-actual}
  \end{minipage}
\end{figure*}

  Since \vecmap{} is pipelined by the initialization and the self-learning, we can assume the failure of unsupervised learning comes from either or both of these two components. Two hypotheses arise:
  \begin{enumerate}
    \item The self-learning cannot succeed even if the initialization is perfect;
    \item The initialization is too bad to kick-off the self-learning even if the self-learning is able to succeed.
  \end{enumerate}

  It is easy to verify the first hypothesis: we start the self-learning with a human-annotated seed dictionary. This way assumes a perfect initialization and thus eliminates the impact of the initialization. \tab{tab:semi-supervised} shows that a small seed dictionary is enough for the self-learning to have a good result. This observation reveals that the self-learning is able to succeed. This left us the second hypothesis, that unsupervised learning fails at the initialization.

  Two natural questions come from the second hypothesis:
  \begin{enumerate}
    \item How to quantify the initialization performance, i.e., the quality of the dictionary generated by the initialization (initial dictionary in short)?
    \item How well the initial dictionary need to be so that the self-learning can succeed?
  \end{enumerate}

  One might expect the accuracy is a good proxy of the quality of a dictionary. But the accuracy only takes the correct translations into account. The intuition is that though the system fails to find the correct translation, it can still be useful if its translations are close to the correct answers. Thus we evaluate the average cosine similarity between the word embeddings of the system translations and the correct answers, dubbed \emph{dictionary similarity}. It will score high if the translations are close to the correct answers.

  \fig{fig:acc-vs-sim} shows how the self-learning performs when starting from dictionaries with different dictionary similarities. These dictionaries are constructed by randomly replacing the translations in the initial dictionary. We can see that the self-learning only succeeds when the dictionary similarities of these starting dictionaries are above some thresholds. These thresholds represent the minimum similarity that the initial dictionary should have for the self-learning to succeed.

  We test the dictionary similarity of the initial dictionary. As shown in \fig{fig:required-vs-actual}, the actual initial dictionary similarities of the initialization are above the thresholds in similar language pairs, but it is the opposite in distant language pairs. The gap between the actual initialization similarity and the minimum similarity for the self-learning to succeed implies the failure of \vecmap{} in distant language pairs.

\section{Proposed Method}
\label{sec:method}

  \begin{table}[t]
    \centering
    \setlength\tabcolsep{5pt}
    \begin{tabular}{l|ccccc|cccc}
      \toprule
      \multicolumn{1}{c|}{Metric} & En & Es & Fr & It & De & Zh & Ja & Vi & Th \\
      \midrule
      Explained variance [\%] & 6.930 & 5.636 & 4.331 & 4.508 & 4.446 & 49.17 & 96.71 & 8.632 & 11.19 \\
      Similarity (before dropmax) & 0.155 & 0.158 & 0.156 & 0.158 & 0.163 & 0.812 & 0.991 & 0.179 & 0.186 \\
      Similarity (after dropmax) & 0.062 & 0.029 & 0.042 & 0.023 & 0.025 & 0.051 & 0.033 & 0.012 & 0.012 \\
      \bottomrule
    \end{tabular}
    \caption{The percentage of variance explained by the highest eigenvalue and the average cosine similarity between any two embeddings before and after applying the dropmax trick.}
    \label{tab:method}
  \end{table}

  \begin{algorithm}[b]
    \caption{Iterative Dimension Reduction}\label{alg:iteration}
    \begin{algorithmic}[1]
      \Procedure{IDR}{$E,n$}\Comment{$E$ is the raw embeddings, $n$ is the initial dimension}
        \State $D \gets \varnothing$\Comment{Set the dictionary to empty}
        \While{$n\le300$}\Comment{300 is the dimension of the raw embeddings}
          \State Reduce $E$ to $\bar{E}$ with dimension $\min(n,300)$ using PCA and dropmax
          \If{$D=\varnothing$}
            \State Run the initialization and the self-learning on $\bar{E}$
          \Else
            \State Run the self-learning on $\bar{E}$ with $D$ as the initial dictionary
          \EndIf
          \State Translate 4K most frequent words and store the results in $D$
          \State $n \gets n \times 2$
        \EndWhile
        \State \textbf{return} $W_X$ and $W_Y$
      \EndProcedure
    \end{algorithmic}
  \end{algorithm}

  As the gap is determined by embeddings and the algorithm, one can improve the algorithm to bridge this gap. In this work, we start with a different angle by simplifying embeddings to make the algorithm easier to succeed. Concretely, we run the dimension reduction on the word embeddings. In principle, the dimension reduction algorithm will drop features that are less important in explaining the data. It can be considered as a way to remove the noise and clean up the data. Here we choose Principal Component Analysis (PCA) \cite{pearson1901} for study.

  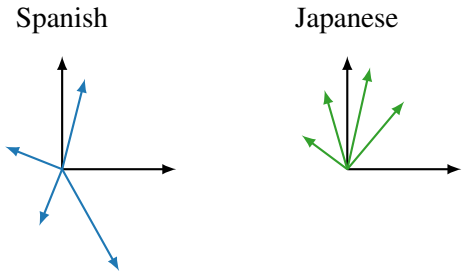
\begin{wrapfigure}{l}{.48\columnwidth}
    \centering
    \begin{tikzpicture}[scale=1.5]
      \coordinate (origin) at (0,0);
      \draw[-latex,thick] (origin) to (1cm,0);
      \draw[-latex,thick] (origin) to (0,1cm);
      \draw[-latex,thick,lyyblue] (origin) to (0.2cm,0.8cm);
      \draw[-latex,thick,lyyblue] (origin) to (-0.5cm,0.2cm);
      \draw[-latex,thick,lyyblue] (origin) to (-0.2cm,-0.5cm);
      \draw[-latex,thick,lyyblue] (origin) to (0.5cm,-0.9cm);
      \node[font=\normalsize,inner sep=0pt,anchor=south] () at ([yshift=1.2cm]origin) {Spanish};

      \begin{scope}[shift={(2.5cm,0)}]
        \coordinate (origin) at (0,0);
        \draw[-latex,thick] (origin) to (1cm,0);
        \draw[-latex,thick] (origin) to (0,1cm);
        \draw[-latex,thick,lyygreen] (origin) to (0.2cm,0.9cm);
        \draw[-latex,thick,lyygreen] (origin) to (-0.2cm,0.7cm);
        \draw[-latex,thick,lyygreen] (origin) to (-0.4cm,0.3cm);
        \draw[-latex,thick,lyygreen] (origin) to (0.5cm,0.6cm);
        \node[font=\normalsize,inner sep=0pt,anchor=south] () at ([yshift=1.2cm]origin) {Japanese};
      \end{scope}
    \end{tikzpicture}
    \caption{An example of embeddings with eigenvectors from PCA (Black arrows are eigenvectors/axes, coloured arrows are projected embedding vectors).}\label{fig:space}
  \end{wrapfigure}

  PCA first computes the covariance matrix of the features in the embeddings. This matrix represents the correlation among features. Then PCA performs the eigenvalue decomposition on this covariance matrix to obtain a set of eigenvectors. Finally, PCA uses eigenvectors with the highest $n$ eigenvalues to project the raw embeddings to space with a lower dimension $n$.

  However, this simple application of PCA only works in a few languages. We find that the highest eigenvalue is much larger than the others in most of the failed languages, e.g., it explains 96\% variance in Japanese while 5.6\% in Spanish as shown in the first row of \tab{tab:method} (the larger the highest eigenvalue is, the more variance it explains). This implies that the embeddings will be stretched the most on the direction of the eigenvector with the highest eigenvalue, resulting in most embeddings pointed to a close direction and being clustered together. \fig{fig:space} shows a simple example of this. The average cosine similarity between every two embeddings in \tab{tab:method} also justifies our hypothesis that the more variance it explains, the closer the embeddings are to each other (higher similarity). Such a phenomenon makes embeddings indistinguishable. The `hubness' problem \cite{DBLP:journals/jmlr/RadovanovicNI10} will occur, where one embedding is the nearest neighbor of many embeddings, even if they have different semantics.

  The simplest solution is that when projecting the embeddings with a dimension $m$ to $n$, we drop not only eigenvectors with the lowest $m-n+1$ eigenvalues but also the one with the highest eigenvalue. We refer to this as the \emph{dropmax} trick. This makes embeddings distinguishable (lower similarity) as shown in the last row of \tab{tab:method} and helps \vecmap{} to succeed in the remaining distant language pairs. Since the dimension reduction incurs the loss of information and hinders the further improvement of our system, we propose \emph{Iterative Dimension Reduction} (IDR), as shown in \alg{alg:iteration}. The algorithm first runs \vecmap{} on embeddings with the smallest dimension $n$ (default set to 50). Then it uses the trained system to translate the $K$ most frequent words (default set to 4K). The resulting dictionary will serve as the initial dictionary for the self-learning in the next step, where it runs on embeddings with a larger dimension (default set to 2$\times$ larger than in the previous step).

  \begin{table}[t!]
    \centering
    \setlength\tabcolsep{5pt}
    \begin{tabular}{l|cccccccc}
      \toprule
      \multicolumn{1}{c|}{System} & En-Es & Es-En & En-Fr & Fr-En & En-It & It-En & En-De & De-En \\
      \midrule
      MUSE \cite{DBLP:conf/iclr/LampleCRDJ18} & 83.20 & 83.66 & 82.66 & 82.39 & 78.20 & 77.90 & 75.10 & 72.93 \\
      \vecmap{} \cite{DBLP:conf/acl/AgirreLA18} & 82.33 & 84.60 & 82.47 & 83.60 & 79.13 & 79.80 & 75.33 & 74.27 \\
      C-MUSE \cite{DBLP:conf/nips/HartmannKS19} & 82.33 & 84.73 & 82.40 & 83.73 & 79.13 & 79.67 & 75.27 & 74.20 \\
      POSTPROC \cite{DBLP:conf/rep4nlp/VulicKG20} & 82.73 & 85.47 & 82.73 & 84.00 & 79.13 & 80.60 & 76.00 & 75.33 \\ 
      \midrule
      Proposed method (dim 50) & 40.33 & 37.40 & 53.53 & 48.07 & 48.13 & 45.00 & 31.60 & 30.33 \\
      Proposed method (dim 100) & 63.47 & 61.80 & 72.73 & 71.13 & 68.40 & 66.73 & 63.67 & 61.27 \\
      Proposed method (dim 200) & 80.33 & 80.27 & 80.40 & 79.67 & 76.47 & 76.67 & 71.27 & 71.20 \\
      Proposed method (dim 300) & 82.40 & 84.60 & 82.60 & 83.67 & 78.93 & 79.67 & 75.33 & 74.33 \\
      \bottomrule
    \end{tabular}
    \caption{The accuracy of different UBDI systems on similar language pairs (the proposed method starts with dimension 50 and doubles the dimension in each step until it reaches 300. We show the performance of each step).}
    \label{tab:similar}
  \end{table}

  \begin{table}[t!]
    \centering
    \setlength\tabcolsep{5pt}
    \begin{tabular}{l|cccccccc}
      \toprule
      \multicolumn{1}{c|}{System} & En-Zh & Zh-En & En-Ja & Ja-En & En-Vi & Vi-En & En-Th & Th-En \\
      \midrule
      MUSE \cite{DBLP:conf/iclr/LampleCRDJ18} & 34.93 & 32.33 & 0.06 & 5.09 & 0 & 0 & 0 & 0 \\
      \vecmap{} \cite{DBLP:conf/acl/AgirreLA18} & 0.07 & 0 & 1.03 & 32.67 & 0.73 & 0.73 & 0 & 0.07 \\
      C-MUSE \cite{DBLP:conf/nips/HartmannKS19} & 39.80 & 34.53 & 0.27 & 32.18 & 0 & 56.8 & 0 & 0 \\
      POSTPROC \cite{DBLP:conf/rep4nlp/VulicKG20} & 0.00 & 44.07 & 0.00 & 34.60 & 0.13 & 0.07 & 0.07 & 0.00 \\ 
      \midrule
      Proposed method (dim 50) & 0.07 & 0 & 10.08 & 6.89 & 0.13 & 0.20 & 0.40 & 0 \\
      Proposed method (dim 100) & 0.2 & 24.47 & 27.55 & 19.99 & 0.87 & 0.40 & 10.40 & 5.87 \\
      Proposed method (dim 200) & 34.13 & 35.93 & 39.07 & 28.67 & 2.60 & 2.13 & 19.47 & 11.75 \\
      Proposed method (dim 300) & 37.33 & 35.27 & 48.87 & 33.08 & 47.6 & 55.53 & 21.60 & 13.64 \\
      \bottomrule
    \end{tabular}
    \caption{The accuracy of different UBDI systems on distant language pairs (the proposed method starts with dimension 50 and doubles the dimension in each step until it reaches 300. We show the performance of each step).}
    \label{tab:distant}
  \end{table}

\section{Experiments}
  
\subsection{Setup}

  We compare our method with four popular UBDI systems: MUSE\footnote{\url{https://github.com/facebookresearch/MUSE}} \cite{DBLP:conf/iclr/LampleCRDJ18}, \vecmap{}\footnote{\url{https://github.com/artetxem/vecmap}} \cite{DBLP:conf/acl/AgirreLA18}, C-MUSE \cite{DBLP:conf/nips/HartmannKS19} and POSTPROC \cite{DBLP:conf/rep4nlp/VulicKG20}. We reproduce the C-MUSE and POSTPROC system using Python. All these systems are run with the default hyper-parameters settings. Our method is based on the open-sourced \vecmap{} implementation.
  
  We evaluate the baseline and our method on 4 similar language pairs, En-\{Es, Fr, It, De\}, and 4 distant language pairs, En-\{Zh, Ja, Vi, Th\}. We use the pretrained 300-dimensional fastText embeddings \cite{DBLP:journals/tacl/BojanowskiGJM17}\footnote{\url{https://fasttext.cc/docs/en/pretrained-vectors.html}}. The evaluation dictionaries are from \newcite{DBLP:conf/iclr/LampleCRDJ18}. We trim all vocabularies to the 20K most frequent words for training. Specifically, \vecmap{} retains the top-4K words for the initialization, while others use the whole vocabulary. All experiments are done on a single Nvidia GTX 1080Ti. We run each experiment 3 times but with different random seeds, then pick the one with the highest cosine similarity of induced nearest neighbors as the final result. This unsupervised model selection criterion has shown to correlate well with UBDI performance \cite{DBLP:conf/nips/HartmannKS19}.

\subsection{Results}

  \tab{tab:similar} shows the results of various systems on similar language pairs, En-\{Es, Fr, It, De\} and their reverse \{Es, Fr, It, De\}-En. We can see that all baseline systems perform well on these language pairs. \vecmap{} and C-MUSE outperform MUSE in most cases. This is because both systems employ the dropout trick in their self-learning processes, which has proven to be effective in jumping out of the local optimum \cite{DBLP:conf/acl/AgirreLA18,DBLP:conf/nips/HartmannKS19}. However, all these baseline systems perform poor on distant language pairs, En-\{Zh, Ja, Vi, Th\} and their reverse \{Zh, Ja, Vi, Th\}-En, as shown in \tab{tab:distant}. C-MUSE is better than the others by obtaining positive results on En-Zh, Zh-En, Ja-En and Vi-En tasks, but still fails on other tasks, i.e., having an accuracy below 5\%.
  
  Our method is based on \vecmap{}, thus it has a good performance in similar language pairs, as shown in the last row of \tab{tab:similar}. On the other hand, our method is robust to distant language pairs as shown in \tab{tab:distant}. In all four distant language pairs and the two directions, our method obtains much better results than the baselines. For example, our method has an accuracy of 21.6\% in En-Th and 13.64\% in Th-En, where none of the baseline systems has an accuracy above 1\%.

  We also observe that the performance of our method in low-dimensional space is much worse than the one in high-dimensional space. For instance, our method has an accuracy of 21.6\% in En-Th when the dimension is 300, while only 10.4\% when the dimension is 100. This observation justifies our previous claim in Section \ref{sec:method} that dimension reduction incurs the loss of information and thus hinders the further improvement of our method. But directly run on high-dimensional embeddings does not succeed, as none of the baselines consistently has an accuracy above 0\% in raw 300-dimensional embeddings. Therefore, it is necessary to run on the low-dimensional space first as a warm start of the high-dimensional counterpart to obtain better performance.
  
\section{Analysis}

\subsection{Ablation Study}

  \begin{table}[t!]
    \centering
    \begin{tabular}{l|cccccccc}
      \toprule
      \multicolumn{1}{c|}{System} & En-Zh & Zh-En & En-Ja & Ja-En & En-Vi & Vi-En & En-Th & Th-En \\
      \midrule
      Proposed method & 37.33 & 35.27 & 48.87 & 33.08 & 47.6 & 55.53 & 21.60 & 13.64 \\
      \midrule
      - IDR & 40.20 & 41.07 & 47.64 & 34.18 & 0.13 & 0.20 & 0.07 & 0 \\
      - Dropmax & 0 & 0 & 0 & 0 & 0.33 & 0.27 & 20.73 & 14.25 \\
      \bottomrule
    \end{tabular}
    \caption{Ablation study of the proposed method on distant language pairs.}
    \label{tab:ablation}
  \end{table}

  \tab{tab:ablation} shows the results of using IDR and dropmax solely on distant language pairs. We can see that IDR is crucial to En-Th and Th-En. In Section \ref{sub-sec:isomorphic}, we show that dimension reduction helps to obtain isomorphic embeddings. These embeddings match the isomorphic assumption made by \vecmap{}.

  On the other hand, the dropmax trick is crucial to En-\{Zh, Ja\} and their reverse. This fact relates well with the observation in \tab{tab:method}, where these two languages suffer from the hubness problem due to the highest eigenvalue. The dropmax trick avoids this issue by removing this highest eigenvalue, as shown in Section \ref{sub-sec:hubness}. We also see that both IDR and dropmax are crucial to En-Vi and Vi-En, which implies that the hubness and isomorphism are their central problems.

\subsection{Isomorphism}
\label{sub-sec:isomorphic}

  Many UBDI methods, including \vecmap{}, make the isomorphic assumption, that the underlying nearest neighbor graphs of two language embedding spaces are connected in the same way. \newcite{DBLP:conf/acl/SogaardVR18} propose the eigenvector similarity to measure how well this assumption is held. Here we are interested in how the isomorphism of the underlying graphs change when the dimension is different. We first normalize, center and normalize the embeddings as in the pre-processing step, calculate the nearest neighbor graphs of the 10K most frequent words in each language, and compute their Laplacian matrices $L_1$ and $L_2$. We then find the smallest $k_1$ such that the sum of the largest $k_1$ eigenvalues of $L_1$ is at least 90\% of the sum of all its eigenvalues, and analogously for $k_2$ and $L_2$. Finally we set $k=\min(k_1,k_2)$, and define the eigenvector similarity of the two graphs as the sum of the squared differences between the $k$ largest eigenvalues $\lambda$ of $L_1$ and $L_2$, $\triangle=\sum_{i=1}^k(\lambda_{1i}-\lambda_{2i})^2$. The higher $\triangle$ is, the less similar the graphs are.

  As shown in \fig{fig:isomorphic}, the eigenvector similarity drops significantly when the dimension is reduced. This implies that the underlying nearest neighbor graphs of two languages become similar in low-dimensional space. This helps the algorithm to succeed in low-dimensional space as the assumption it makes is held. This phenomenon might be the result that many language pairs share some principle axes of variation, especially the ones with high eigenvalues \cite{DBLP:conf/emnlp/HoshenW18}.

  \begin{figure*}[t]
    \begin{minipage}[b][][b]{.48\columnwidth}
      \centering
      \begin{tikzpicture}
        \begin{axis}[
          width=\linewidth,height=5cm,
          enlarge x limits=0.05,
          enlarge y limits={upper,value=0.35},
          ymin=100,ymax=700,
          ylabel={Eigenvector Similarity $\triangle$},
          ylabel near ticks,
          xlabel={Dimension},
          xlabel near ticks,
          xmajorgrids=true,
          ymajorgrids=true,
          grid style=dashed,
          xtick=data,
          every tick label/.append style={font=\small},
          label style={font=\small},
          legend style={font=\small,at={(0.5,0.98)},anchor=north,inner sep=3pt},
          legend image post style={scale=0.8},
          legend columns=4,
          legend cell align={left},
        ]
        \addplot [lyyblue,mark=*] coordinates {
          (50,126) (100,395) (200,624) (300,610)
        };\addlegendentry{En-Zh}
        \addplot [lyyred,mark=square*] coordinates {
          (50,324) (100,488) (200,681) (300,674)
        };\addlegendentry{En-Ja}
        \addplot [lyygreen,mark=triangle*] coordinates {
          (50,422) (100,503) (200,520) (300,484)
        };\addlegendentry{En-Vi}
        \addplot [orange,mark=diamond*] coordinates {
          (50,505) (100,606) (200,676) (300,647)
        };\addlegendentry{En-Th}
        \end{axis}
      \end{tikzpicture}
      \captionof{figure}{The eigenvector similarity vs. different dimensions on distant language pairs.}
      \label{fig:isomorphic}
    \end{minipage}
    \hfill
    \begin{minipage}[b][][b]{.48\columnwidth}
      \centering
      \begin{tikzpicture}
        \begin{axis}[
          width=\linewidth,height=5cm,
          symbolic x coords={En-Es,En-Fr,En-It,En-De,En-Zh,En-Ja,En-Vi,En-Th},
          ymin=25,
          enlarge x limits=0.1,
          enlarge y limits={upper,value=0.8},
          ylabel={Hubness $H$},
          ylabel near ticks,
          ybar=0pt,
          xtick=data,
          ytick=\empty,
          nodes near coords,
          every node near coord/.append style={rotate=90,anchor=west,font=\scriptsize},
          every tick label/.append style={font=\footnotesize},
          xticklabel style={rotate=45,anchor=north east,font=\footnotesize,inner sep=0pt,outer sep=2pt},
          bar width=5.5pt,
          xmajorgrids=true,
          ymajorgrids=true,
          grid style=dashed,
          label style={font=\small},
          legend image code/.code={
              \draw [#1] (0cm,-0.1cm) rectangle (0.3cm,0.1cm);
          },
          legend style={font=\small,at={(0.5,0.98)},anchor=north,inner sep=3pt,column sep=3pt},
          legend cell align={left},
          legend columns=2,
        ]
          \addplot [draw=lyyblue!80,fill=lyyblue!60,pattern=north west lines,pattern color=lyyblue] coordinates {(En-Es,57.26) (En-Fr,60.49) (En-It,56.33) (En-De,50.39) (En-Zh,32.87) (En-Ja,37.86) (En-Vi,40.04) (En-Th,49.93)};\addlegendentry{Before dropmax}
          \addplot [draw=lyygreen!80,fill=lyygreen!60,pattern=north east lines,pattern color=lyygreen] coordinates {(En-Es,57.48) (En-Fr,60.76) (En-It,56.73) (En-De,50.81) (En-Zh,41.97) (En-Ja,45.46) (En-Vi,40.08) (En-Th,54.66)};\addlegendentry{After dropmax}
        \end{axis}
      \end{tikzpicture}
      \captionof{figure}{The hubness level before and after applying the dropmax trick.}
      \label{fig:hubness}
    \end{minipage}
  \end{figure*}

\subsection{Hubness}
\label{sub-sec:hubness}

  Cross-lingual word embeddings are known to suffer from the hubness problem \cite{DBLP:conf/iclr/LampleCRDJ18}, where a few points (known as \emph{hubs}) are the nearest neighbors of many other points in high-dimensional spaces. As suggested in Section \ref{sec:method}, distant language pairs might suffer more from this problem and the dropmax trick helps to alleviate this problem. Thus we would like to know to what extent the dropmax trick helps in the hubness problem. Here we adopt the hubness metric proposed by \newcite{DBLP:conf/acl/OrmazabalALSA19} for evaluation. This metric measures the percentage of target words $H$ that are the nearest neighbor of all the source words. For instance, a hubness value of $H = 60\%$ would indicate that 60\% of the target words are the nearest neighbors of all the source words. This way, lower values of $H$ are indicative of a higher level of hubness.

  \fig{fig:hubness} is the hubness level of different languages before and after applying the dropmax trick. We can see that the dropmax trick generally alleviates the hubness problem, even when the hubness level is low. For example, En-It has $H=56.39\%$ before applying the dropmax trick. After that, it has $H=56.73\%$, a $0.34\%$ improvement on the hubness level. For those language pairs with a high hubness level such as En-Zh and En-Ja, the improvement is obvious, e.g., more than 9\% improvement on En-Zh.

\subsection{Dictionary Similarity}

  As suggested in Section \ref{sec:method}, dimension reduction helps to bridge the performance gap and we examine it here. This gap can be measured by the dictionary similarity proposed in Section \ref{sec:analysis}. Here we choose En-\{Vi, Th\} for study, since dimension reduction is crucial to their successes as shown in \tab{tab:ablation}.

  In \fig{fig:sim-vs-dim}, we can see that the gap between the initialization and the self-learning varies in different dimensions. In the dimension that \vecmap{} succeeds, the actual initial dictionary similarity is much higher than the minimum initial dictionary similarity to succeed, e.g., 0.58 vs. 0.15 in 300 dimensions for En-Vi and 0.51 vs. 0.15 in 100 dimensions for En-Th. This means that \vecmap{} in the previous dimension reduction step generates a good initial dictionary. When looking at the results in the previous step, we find that the actual initial dictionary similarity is close to but not surpass the minimum initial dictionary similarity to succeed, e.g., 0.15 vs. 0.17 in 200 dimensions for En-Vi and 0.15 vs. 0.2 in 50 dimensions for En-Th. This implies that though \vecmap{} does not succeed in that dimension, closing this gap already allows it to translate well on the frequent words, i.e., words to construct the initial dictionary. This enables unsupervised learning in the next dimension reduction step.

  \begin{figure*}
    \begin{minipage}[b][][b]{.48\columnwidth}
      \hspace*{\fill}
      \begin{tikzpicture}
        \begin{axis}[
          width=0.5\linewidth,height=5cm,
          enlarge x limits=0.2,
          enlarge y limits={upper,value=0.5},
          ylabel={Dictionary Similarity},
          ylabel near ticks,
          xlabel={Dimension (En-Vi)},
          xlabel near ticks,
          xmajorgrids=true,
          ymajorgrids=true,
          grid style=dashed,
          xtick=data,
          ytick=\empty,
          symbolic x coords={50,100,200,300},
          ymin=0,
          ybar stacked,
          every tick label/.append style={font=\small},
          nodes near coords,
          every node near coord/.append style={text=black,font=\scriptsize},
          label style={font=\small},
          legend image code/.code={
              \draw [#1] (0cm,-0.1cm) rectangle (0.3cm,0.1cm);
          },
          legend style={font=\small,at={(0.5,0.98)},anchor=north,inner sep=2pt,column sep=3pt},
          legend cell align={left},
          legend columns=1,
        ]
        \addplot+ [draw=lyyblue!80,fill=lyyblue!60,pattern=north west lines,pattern color=lyyblue,ybar] coordinates {
          (50,0.4265) (100,0.2569) (200,0.1692) (300,0.1478)
        };\addlegendentry{Threshold}
        \addplot+ [draw=lyygreen!80,fill=lyygreen!60,pattern=north east lines,pattern color=lyygreen,ybar] coordinates {
          (50,0.2454) (100,0.1759) (200,0.1537) (300,0.5824)
        };\addlegendentry{Actual}
        \end{axis}
      \end{tikzpicture}
      \hfill
      \begin{tikzpicture}
        \begin{axis}[
          width=0.5\linewidth,height=5cm,
          enlarge x limits=0.2,
          enlarge y limits={upper,value=0.5},
          ylabel={Dictionary Similarity},
          ylabel near ticks,
          xlabel={Dimension (En-Th)},
          xlabel near ticks,
          xmajorgrids=true,
          ymajorgrids=true,
          grid style=dashed,
          xtick=data,
          ytick=\empty,
          symbolic x coords={50,100,200,300},
          ymin=0,
          ybar stacked,
          every tick label/.append style={font=\small},
          nodes near coords,
          every node near coord/.append style={text=black,font=\scriptsize},
          label style={font=\small},
          legend image code/.code={
              \draw [#1] (0cm,-0.1cm) rectangle (0.3cm,0.1cm);
          },
          legend style={font=\small,at={(0.5,0.98)},anchor=north,inner sep=2pt,column sep=3pt},
          legend cell align={left},
          legend columns=1,
        ]
        \addplot+ [draw=lyyblue!80,fill=lyyblue!60,pattern=north west lines,pattern color=lyyblue,ybar] coordinates {
          (50,0.1981) (100,0.1483) (200,0.1205) (300,0.1865)
        };\addlegendentry{Threshold}
        \addplot+ [draw=lyygreen!80,fill=lyygreen!60,pattern=north east lines,pattern color=lyygreen,ybar] coordinates {
          (50,0.1536) (100,0.5088) (200,0.5834) (300,0.6152)
        };\addlegendentry{Actual}
        \end{axis}
      \end{tikzpicture}
      \hspace*{\fill}
      \captionof{figure}{The minimum initial dictionary similarity to succeed (Threshold) and the actual initial dictionary similarity (Actual) vs. dimensions.}
      \label{fig:sim-vs-dim}
    \end{minipage}
    \hfill
    \begin{minipage}[b][][b]{.48\columnwidth}
      \begin{tikzpicture}
        \begin{axis}[
          width=0.49\linewidth,height=5cm,
          yticklabel style={/pgf/number format/fixed},
          xlabel={Oracle Structural Sim.},
          ylabel={Initial Acc. [\%]},
          xlabel near ticks,
          enlarge x limits=0.3,
          enlarge y limits=0.2,
          xmajorgrids=true,
          ymajorgrids=true,
          grid style=dashed,
          every tick label/.append style={font=\small},
          label style={font=\small},
          every node near coord/.append style={font=\small},
          xlabel style={yshift=-0.06cm},
        ]
        \addplot [
          scatter/classes={a={lyyblue}, b={lyyred}},
          scatter,only marks,mark=*,
          scatter src=explicit symbolic,
          nodes near coords*={\label},
          every node near coord/.append style={anchor=\anchor},
          visualization depends on={value \thisrow{label}\as\label},
          visualization depends on={value \thisrow{anchor}\as\anchor},
        ] table [meta=class] {
          x y label class anchor
          0.2943 1.51 En-Es b {east}
          0.3039 1.40 En-Fr b {north}
          0.2748 1.14 En-It b {east}
          0.3283 0.98 En-De b {north}
          0.2400 0.13 En-Zh a {north west}
          0.2220 0.13 En-Ja a {south}
          0.2078 0.09 En-Vi a {north}
          0.2503 0.22 En-Th a {west}
        };
        \end{axis}
      \end{tikzpicture}
      \hfill
      \begin{tikzpicture}
        \begin{axis}[
          width=0.49\linewidth,height=5cm,
          yticklabel style={/pgf/number format/fixed},
          xlabel={Maximum Acc. [\%]},
          ylabel={Initial Acc. [\%]},
          xlabel near ticks,
          enlarge x limits=0.3,
          enlarge y limits=0.2,
          xmajorgrids=true,
          ymajorgrids=true,
          grid style=dashed,
          every tick label/.append style={font=\small},
          label style={font=\small},
          every node near coord/.append style={font=\small},
          xlabel style={yshift=-0.06cm},
        ]
        \addplot [
          scatter/classes={a={lyyblue}, b={lyyred}},
          scatter,only marks,mark=*,
          scatter src=explicit symbolic,
          nodes near coords*={\label},
          every node near coord/.append style={anchor=\anchor},
          visualization depends on={value \thisrow{label}\as\label},
          visualization depends on={value \thisrow{anchor}\as\anchor},
        ] table [meta=class] {
          x y label class anchor
          68.26 1.51 En-Es b {south}
          66.26 1.40 En-Fr b {east}
          61.06 1.14 En-It b {east}
          66.88 0.98 En-De b {north}
          34.67 0.13 En-Zh a {south}
          33.54 0.13 En-Ja a {north}
          39.59 0.09 En-Vi a {west}
          42.77 0.22 En-Th a {south west}
        };
        \end{axis}
      \end{tikzpicture}
      \captionof{figure}{The initial accuracy (Initial Acc.) vs. the oracle structural similarity (Oracle Structural Sim.) and the maximum accuracy (Maximum Acc.).}
      \label{fig:init-acc}
    \end{minipage}
  \end{figure*}

\subsection{Understanding the Initialization}

  In \fig{fig:required-vs-actual}, the initialization is poor on distant language pairs as indicated by the gap. We investigate which factor might have an impact on the accuracy of the initial dictionary (initial accuracy in short). It allows us to identify the obstacles in the initialization of distant language pairs. Here we measure the initial accuracy instead of the proposed dictionary similarity, as its value depends on embeddings and thus can not be compared across languages.

  Here we study two factors: the oracle structural similarity and the maximum accuracy. Structural similarity is the cosine similarity between rows of $\sqrt{M_X}$ and $\sqrt{M_Y}$, where a row in $\sqrt{M_X}$ associates to a source word and represents the similarities between this source word and other source words, and analogously for $\sqrt{M_Y}$. This similarity is used in the initialization to select word translations to construct the initial dictionary. The oracle structural similarity is the average of the structural similarity of all possible and correct word translations in the initialization. This oracle structural similarity measures how strong the assumption made in the initialization is, which assumes aligned source and target words should have high structural similarity. The higher the oracle structural similarity is, the easier the initialization finds correct translations. As shown in the left part of \fig{fig:init-acc}, the oracle structural similarity is positively related to the initial accuracy. Distant language pairs have a low similarity, which means that the assumption made in the initialization is unlikely to hold.

  The maximum accuracy is the accuracy that is obtained by a perfect initialization strategy. It is lower than 100\% as the vocabulary in the initialization might not contain the translations of some source and target words. We see that in the right part of \fig{fig:init-acc}, the maximum accuracy is also positively related to the initial accuracy. Distant language pairs have a low maximum accuracy, which means that the initialization is less likely to find the correct translation for a source word.

\subsection{Error Analysis}

  In \tab{tab:distant}, we can see that even for our best system it still has low accuracy in distant language pairs. To identify the main source of errors, we perform an error analysis of the system output in a sized 5K En-Zh dictionary from \newcite{DBLP:conf/iclr/LampleCRDJ18}. We randomly sample 200 error examples and let a human expert to classify these examples into four main categories: the answer and the translation are correct (CC), the answer is correct and the translation is wrong (CW), the answer is wrong and the translation is correct (WC), the answer and the translation are wrong (WW).

  In our analysis, there are 25.5\% errors are CC. This is due to the polysemy of words and the dictionary does not cover all possible translations. A few cases are WC (0.5\%) and WW (3\%). This means that there are some minor issues on the dictionary quality. For the main category CW (71\%), there are 17\% resulted from proper nouns, which have shown to be meaningless in the evaluation \cite{DBLP:conf/emnlp/Kementchedjhieva19}. 18.5\% have a close meaning to the answer. 10.5\% are the untranslated error, where the translation is identical to the source word. The remaining 25\% are true errors, e.g., antonym.

\section{Related Work}

  In recent years a number of methods have been proposed to learn bilingual dictionary from monolingual word embeddings. Early work \cite{DBLP:journals/corr/MikolovLS13} relies on a seed dictionary to learn the source-target word embedding mapping. \newcite{DBLP:conf/naacl/XingWLL15} enforce the word embeddings to be of unit length and the orthogonal constraint on the linear mapping. \newcite{DBLP:conf/eacl/FaruquiD14} on the other hand use Canonical Correlation Analysis (CCA) to project both source and target embeddings to a common low-dimensional space. \newcite{DBLP:conf/emnlp/ArtetxeLA16} show that the above methods are variants of the same objective. \newcite{DBLP:conf/iclr/SmithTHH17} further show that this objective is closely related to the orthogonal Procrustes problem. \newcite{DBLP:conf/acl/ArtetxeLA17} obtain competitive results using the self-learning with a seed dictionary of only 25 word pairs.

  \noindent\textbf{Adversarial methods.} \newcite{DBLP:conf/acl/ZhangLLS17} attempt the unsupervised bilingual dictionary induction task using the adversarial network. They use a generator to transform the source word embeddings to the target word embeddings and a discriminator to classify whether the given embedding is sampled from the true target word embeddings or generated by the generator. The generator is trained to fool the discriminator and the discriminator is trained to identify the generated word embeddings. In the end, the generator will be used to induce the bilingual dictionary. Their following work \cite{DBLP:conf/emnlp/ZhangLLS17} minimizes Earth-Mover's distance between the transformed source and target embeddings distribution. \newcite{DBLP:conf/iclr/LampleCRDJ18} improve the results by treating the dictionary produced by the adversarial network as the seed dictionary of the self-learning. To mitigate the hubness problem \cite{DBLP:journals/jmlr/RadovanovicNI10}, they propose an effective nearest neighbors retrieval method CSLS for dictionary induction. \newcite{DBLP:conf/emnlp/XuYOW18} minimize Sinkhorn distance instead and introduce the circle consistency such that a source word embedding can be translated back after translating it to a target word. \newcite{DBLP:conf/naacl/MohiuddinJ19} extract latent codes from word embeddings and align words according to their latent codes.

  \noindent\textbf{Non-adversarial methods.} There is another line of research that focuses on a non-adversarial approach. \newcite{DBLP:conf/acl/AgirreLA18} propose a heuristic to induce an initial dictionary by exploiting the structural similarity of embeddings. They also propose the stochastic dictionary induction method, which significantly improves the robustness as well as the performance of self-learning. \newcite{DBLP:conf/emnlp/HoshenW18} assume that many language pairs share some principle axes of variation. Therefore they first use PCA to project the word embeddings to a lower-dimensional space. Then they apply a variant of the Iterative Closest Point method to find the source and target word embeddings mapping. \newcite{DBLP:conf/naacl/ZhouMWN19} use normalizing flows to match the distribution of source and target word embeddings. But they rely on a numeral seed dictionary and the additional word frequency information. More recently, \newcite{DBLP:conf/nips/HartmannKS19} find that more robust results can be obtained by using the adversarial method to produce the initial dictionary for the advanced self-learning (with the stochastic dictionary induction). \newcite{DBLP:conf/acl/ArtetxeLA19a} first generate a pseudo parallel corpus by an unsupervised machine translation system. They then extract a bilingual dictionary from the word alignment learned on that corpus. This simple process shows much better results than previous methods. \newcite{DBLP:conf/rep4nlp/VulicKG20} introduce a simple post-processing step to improve UBDI performance on distant language pairs.


\section{Conclusion}

  In this work, we pinpoint in which part the representative UBDI system, \vecmap{}, fails on distant language pairs. We identify a gap between the initialization performance and the minimum initialization performance for the self-learning to succeed, which is responsible for its failure. We propose Iterative Dimension Reduction to bridge this gap. Our method obtains substantial gains in distant language pairs without scarifying the performance of similar language pairs. It has shown to robust to the four distant language pairs we experiment with.

\section*{Acknowledgement}

  This work was supported in part by the National Science Foundation of China (Nos. 61876035 and 61732005), the National Key R\&D Program of China (No. 2019QY1801). The authors would also like to thank anonymous reviewers for their valuable comments.

\bibliographystyle{coling}
\bibliography{coling2020}

\end{document}